%% file: main.tex
\documentclass{article}

\usepackage{microtype}
\usepackage{graphicx}
\usepackage{subfigure}
\usepackage{booktabs} 

\usepackage{hyperref}

\usepackage[accepted]{icml2024}

\usepackage{amsmath}
\usepackage{amssymb}
\usepackage{mathtools}
\usepackage{amsthm}

\usepackage[capitalize,noabbrev]{cleveref}

\theoremstyle{plain}

\theoremstyle{definition}

\theoremstyle{remark}

\usepackage[textsize=tiny]{todonotes}
\usepackage{enumitem}
\usepackage{bm}
\usepackage{multirow}
\usepackage{array}

\icmltitlerunning{\method{}: Continuous Spiking Graph Neural Networks}

\begin{document}
\def\method{GG-DAOD}
\twocolumn[
\icmltitle{Utilizing Graph Generation for Enhanced Domain Adaptive Object Detection}



\icmlsetsymbol{equal}{*}

\begin{icmlauthorlist}
\icmlauthor{Mu Wang}{}
\end{icmlauthorlist}



\icmlkeywords{Machine Learning, ICML}

\vskip 0.3in
]




\input{code/0_abstract}

\input{code/1_introduction}
\input{code/2_related}
\input{code/4_method}
\input{code/5_exp}
\input{code/6_conclusion}

\bibliography{main}
\bibliographystyle{icml2024}

\clearpage

\end{document}

%% file: code/0_abstract.tex
\begin{abstract}
The problem of Domain Adaptive in the field of Object Detection involves the transfer of object detection models from labeled source domains to unannotated target domains. Recent advancements in this field aim to address domain discrepancies by aligning pixel-pairs across domains within a non-Euclidean graphical space, thereby minimizing semantic distribution variance. Despite their remarkable achievements, these methods often use coarse semantic representations to model graphs, mainly due to ignoring non-informative elements and failing to focus on precise semantic alignment. Additionally, the generation of coarse graphs inherently introduces abnormal nodes, posing challenges and potentially biasing domain adaptation outcomes. Consequently, we propose a framework, which utilizes the Graph Generation to enhance the quality of DAOD (\method{}). Specifically, we introduce a Node Refinement module that utilizes a memory bank to reconstruct noisy sampled nodes while applying contrastive regularization to noisy features. To enhance semantic alignment, we propose separating domain-specific styles from category invariance encoded within graph covariances, which allows us to selectively remove domain-specific styles while preserving category-invariant information, thus facilitating more accurate semantic alignment across different domains. Furthermore, we propose a Graph Optimization adaptor, leveraging variational inference to mitigate the impact of abnormal nodes. Extensive experimentation across three adaptation benchmarks validates that \method{} achieves state-of-the-art performance in the task of unsupervised domain adaptation.

\end{abstract}

%% file: code/1_introduction.tex
\section{Introduction}

The advancements in generic object detection methodologies \cite{girshickICCV15fastrcnn,redmon2018yolov3,tian2019fcos} have been impressive, showcasing consistent performance across both training and test datasets. These methodologies are crucial in a multitude of real-world applications, with automated driving being just one example. However, their application in new domains often faces considerable challenges due to domain gaps \cite{chen2018domain}, leading to notable performance degradation. These discrepancies pose limitations on the transferability of object detection models, hindering their effectiveness in diverse contexts.

To address these challenges, considerable research efforts have been directed towards Unsupervised Domain Adaptation (UDA) techniques, aiming to facilitate the transfer of object detectors from annotated source domains to unlabeled target domains. Previous studies \cite{tian2021knowledge,xu2020cross} have explored the semantic space at the category level, employing methods to model semantic information through category centroids. Recent breakthroughs \cite{li2022scan++,shou2023comprehensive,vs2023instance,yu2022mttrans} in Domain Adaptive Object Detection (DAOD) have showcased significant progress, utilizing graph-based methodologies to align pixel-pairs across domains within a non-Euclidean graphical space.
Beyond automated driving, the practical applications of these techniques span a wide array of fields. They are indispensable in surveillance systems for enhancing security measures, in industrial automation for ensuring quality control, in medical imaging for providing diagnostic assistance, and even in retail for optimizing inventory management and analyzing customer behavior. The versatility of adapting object detection models to diverse domains ensures their applicability and efficacy across various real-world scenarios, ultimately fostering improved safety, efficiency, and decision-making processes across numerous industries.

Despite their commendable performance, existing category-level approaches based on graphs \cite{chen2022relation,li2022scan++,ju2024survey,yin2024dynamic,yin2024continuous} face two significant challenges. Firstly, these methods often rely on a coarse semantic sampling strategy, typically applied to ground-truth boxes, to construct graphs. However, this approach introduces bias in domain adaptation, as it may overlook crucial semantic information present outside the annotated bounding boxes. Consequently, the model's ability to adapt to new domains and generalize across different datasets may be hindered. This bias stems from the spatially uniform sampling method employed. Ground truth boxes, commonly used in the source domain, may encompass non-informative background regions irrelevant to the object detection task. Conversely, target maps sampled from pseudo score maps in the target domain may introduce considerable noise due to misclassified labels or inaccuracies in the score estimation process. This noise can adversely affect the adaptation process by providing erroneous or misleading information to the model, leading to suboptimal performance in the target domain. Existing sampling methods rely on these boxes without effectively addressing the non-informative noise. Aligning noise across domains from training batches presents challenges in constructing an implicit model that accurately captures the category-specific distribution. This difficulty can lead to suboptimal semantic alignment and hinder the model's ability to generalize well to the target domain. Moreover, the presence of noise can cause the model to converge quickly to a locally optimal state during stochastic gradient optimization, further exacerbating the problem.
Therefore, adopting a more promising graph modeling strategy becomes imperative to represent the domain-level semantic space in an unbiased and robust manner. Such a strategy should aim to mitigate the impact of noise and effectively capture the underlying semantic structure across domains, thereby facilitating more accurate and reliable adaptation in object detection tasks.

The second challenge pertains to the presence of abnormal nodes within the generated graph, encompassing both nonsensical nodes arising from unclear foreground semantics and out-of-distribution (OOD) nodes sampled from unambiguous foreground pixels during the graph generation process. Sampling methods reliant on ambiguous semantics within ground truth often yield abnormal nodes, including blurred pixels depicting objects blocking each other in extreme weather conditions. These abnormal nodes introduce noise and ambiguity into the graph representation, complicating the process of semantic alignment across domains. Consequently, the performance of object detection models may be adversely affected, leading to suboptimal results, particularly in challenging scenarios such as extreme weather conditions. Thus, it is crucial to develop more robust sampling methods capable of effectively handling ambiguous semantics and reducing the presence of abnormal nodes in the graph representation to enhance overall performance.
The current strategy assumes that both training and testing data are sampled from the same distribution (ID) across graphs, which introduces uncertainty in Domain Adaptive Object Detection (DAOD) tasks on graphs. Unlike Euclidean data, where DAOD tasks aim to predict each node, the interconnected nature of nodes within the graph structure leads to non-identical distributions in node generation. Odd nodes may arise from pixels distant from the object's center, overlapped objects, or blurred objects. Confusion with these abnormal nodes can lead to misclassifications, especially severe when the model predicts false pseudo-labels in the target domain, resulting in erroneous semantic node sampling. Graph completion methods, as employed in \cite{chen2022relation,li2022scan++,li2022sigma}, are typically derived from ground truth data. However, these methods often introduce various abnormal nodes with peculiar semantics due to the coarse nature of the data and the inherent features of the graph generation process. Aligning odd nodes with others undermines the model's ability to represent category information, potentially leading to significant missteps during optimization \cite{ren2018learning}, or overfitting. Identifying these odd nodes proves challenging, particularly in graph data containing structural knowledge, thwarting traditional outlier detection methods \cite{li2022graphde}. These observations motivate the development of an intuitive approach for down-weighting abnormal nodes in Graph Optimization.

To address the issues mentioned above, we introduce a Graph Generation framework designed to model unbiased semantic graphs. This framework aims to tackle non-informative noise and identify abnormal nodes within generative graphs, thereby enhancing domain adaptation. Illustrated in Figure 1, drawing inspiration from contrastive learning methodologies \cite{chen2020simple,li2022sigma}, we introduce a Node Refinement module to rectify mismatched semantics, generating hallucination nodes for missing categories while mitigating non-informative noise. Specifically, we reconstruct raw node features based on categorical relations with absent nodes from the Graph Memory Bank module. To attain precise semantic alignment across domains, we decouple graph covariance to remove domain-specific elements, thus capturing essential invariant characteristics of each category. Addressing the second issue, we introduce a Graph Optimization adaptor to filter out abnormal nodes. In summary, the contributions are summarized as follows.

\begin{itemize}
    \item We introduce a Graph Generation framework tailored for DAOD, offering a novel approach to graph modeling with comprehensive semantics aimed at effectively aligning class-conditional distributions.
    \item The Node Refinement module is designed to tackle missing semantics while mitigating non-informative noise. To improve semantic alignment, we disentangle domain-specific style from category-invariant content encoded within graph covariance and selectively remove domain-specific elements.
    \item To enhance the quality of the graph generation, we introduce a Graph Optimization adaptor, which infers the environment latent variable to down-weight abnormal nodes harboring peculiar semantics.
    \item Extensive experiments validate that our proposed framework achieves state-of-the-art performance across various datasets, significantly surpassing existing counterparts.
\end{itemize}

%% file: code/2_related.tex
\section{Related Work}
\subsection{Object Detection}
The initial methods of object detection relied on sliding window techniques followed by classifiers utilizing hand-crafted features \cite{dalal2005histograms,felzenszwalb2009object,viola2001rapid}. With the emergence of deep convolutional neural networks (CNNs), methods such as R-CNN \cite{girshick2014rich}, SPPNet \cite{he2015spatial}, and Fast R-CNN \cite{girshickICCV15fastrcnn} have been introduced. These methods leverage CNNs to extract image features and use the fast region proposal instead of the slower sliding window~\cite{uijlings2013selective}. 
Recent advancements in the area of object detection have significantly improved bounding box detection. For example, Faster R-CNN \cite{girshick2015fast} introduced a region proposal network (RPN) to predict refinements in the locations and sizes of predefined anchor boxes. Similarly, SSD \cite{liu2016ssd} performs classification and bounding box prediction directly on feature maps at multiple scales using anchor boxes with different aspect ratios.
YOLO \cite{redmon2016you} tackles a regression problem on a grid, predicting the bounding box and class label for each cell in the grid containing the center of an object. Further extensions are explored in \cite{ZhangLLH16,RedmonF17,DaiLHS16}.  Accordingly, we build upon Faster R-CNN, a slower but more accurate detector, to propose our improvements.

\subsection{Domain Adaptation}
Domain adaptation initially received extensive attention in the context of image classification, with a predominant focus in the domain adaptation literature \cite{zhang2022would,zhang2023investigating,meng2024deep,anonymous2024dream,yin2024dream,yin2022deal,yin2023coco}. Various methods have been developed to address this problem, such as cross-domain kernel approaches\cite{DuanXTL12,DuanTX12,GongSSG12}, and numerous strategies aim to obtain domain-invariant prediction\cite{KulisSD11,GopalanLC11,FernandoHST13,SunFS16,WangLDG17,yin2023}.
The advent of deep learning has spurred advancements in domain-invariant feature learning. In \cite{LongC0J15,LongZ0J16}, a method is proposed that learns a reproducing kernel Hilbert embedding of hidden features within the network and conducts mean-embedding matching for both domain distributions. Similarly, \cite{GaninL15,GaninUAGLLML16} utilize an adversarial loss coupled with a domain classifier to learn discriminative and domain-invariant features. 
However, domain adaptation for object detection has garnered less attention. 
Notable works in domain adaptation for non-image classification tasks include \cite{GebruHF17,Chen0G18,KhodabandehJZP18,abs-1805-08916}. Regarding object detection, approaches such as \cite{XuRVL14} utilized an adaptive SVM to mitigate domain shift, \cite{RajNT15} performed subspace alignment on features extracted from R-CNN, and \cite{Chen0SDG18} employed Faster R-CNN as a baseline while adopting an adversarial approach akin to \cite{GaninL15} to jointly learn domain-invariant features across target and source domains.
We propose a novel framework by reframing the problem as noisy labeling. Our approach involves designing a robust training scheme for object detection, where the model is trained on noisy bounding boxes and labels obtained from the target domain, which serve as pseudo-ground truth. This innovative approach aims to tackle domain adaptation challenges in object detection from a unique perspective.

\subsection{Out of Distribution}
Out-of-distribution generalization holds significant importance within the machine learning (ML) community \cite{ai2023gcn,shou2023adversarial,tang2024merging,mansour2009domain}. Previous studies \cite{arjovsky2019invariant} have explored potential environmental latent variables as causes of distribution shifts. Several algorithms have been developed to learn environmental invariant models \cite{creager2021environment,koyama2020invariance}. However, research investigating this problem in the context of graphs remains scarce. Recent works \cite{li2022graphde,wu2022handling} have addressed the out-of-distribution issue on graphs and proposed frameworks for learning invariance. Our work uniquely addresses the out-of-distribution problem on graphs within the domain of Domain Adaptive Object Detection (DAOD) and leverages the Graph Optimization adaptor to mitigate odd semantic nodes.

%% file: code/4_method.tex
\section{The Proposed \method{}}

\begin{figure*}[t]
  \centering
  \includegraphics[scale=0.46]{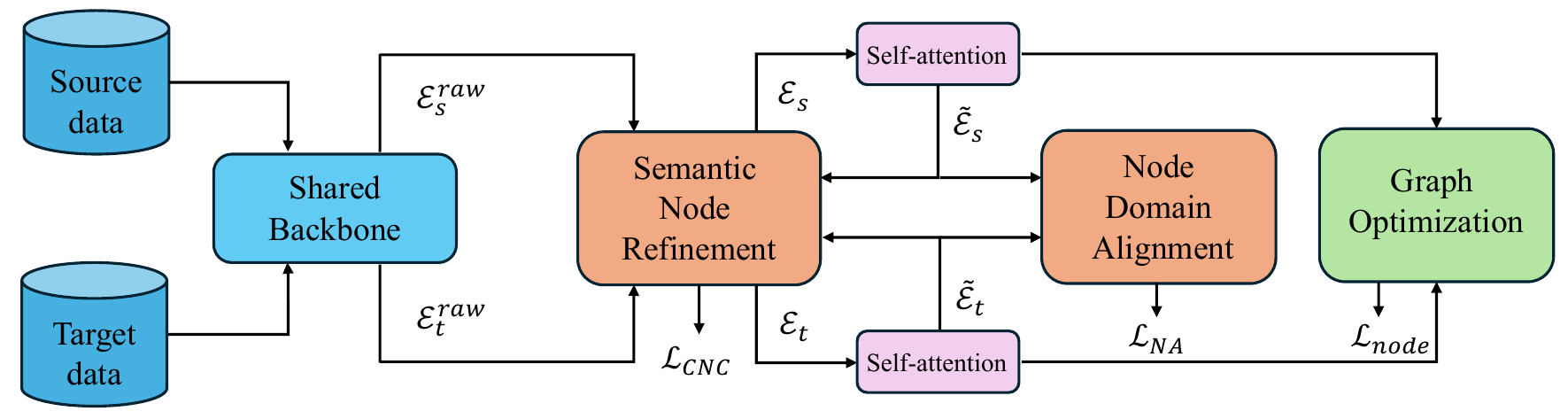}
  \caption{The framework of the proposed \method{}.}
  \label{fig1}
\end{figure*}

The proposed framework of \method{} is illustrated in Figure 1. Beginning with the labeled source images ${(x_s^i,y_s^i)}_{i=1}^{N_s}$ and unlabeled target images $\{x_t^i\}_{i=1}^{N_t}$, each belonging to $C$ classes, where $N_s$ and $N_t$ denote the number of source and target images. A shared feature extractor $\phi$ is utilized to extract image features, which are subsequently inputted into the Node Refinement module. Within this module, visual features undergo an initial transformation into the graphical space facilitated by a Multi-Layer Perceptron (MLP). Following this, the Contrastive Node Completion (CNC) procedure is employed to address semantic mismatches and mitigate non-informative noise, yielding node sets $\mathcal{V}_{s/t}$. This process utilizes contrastive learning to align the initial and the post-semantic distribution. The node set $\mathcal{V}_{s/t}$ is then fed through a shared attention layer to derive highly reliable nodes $\tilde{\mathcal{V}}_{s,t}$. Further enhancing category-level semantics, the Node Covariance Alignment adaptor is engaged to separate domain-specific styles from class-invariant content encoded in node covariance, selectively removing domain-specific elements. Subsequently, the refined nodes $\tilde{\mathcal{V}}_{s,t}$ proceed to the Graph Optimization adaptor. Following the culmination of our refined graph generation process, our model achieves significant adaptation.

\subsection{Semantic Sampling Transformation}
For the source data, we uniform sampling to collect pixels residing within ground-truth boxes, treating them as class-aware foreground nodes, while other pixels are sampled as background nodes. Conversely, for the target data, we generate pseudo scores $\hat{p}_t$ by propagating target features through the classification head. Subsequently, we select pixels satisfying $\max_C (\hat{p}_t^i)>\tau$ as class-aware foreground nodes, while pixels with low scores meeting the $\max_C(\hat{p}_t^i)<\tau$ criteria are chosen as background samples.
To transform visual space into graphical space, we apply a nonlinear projection to the sampled semantic patterns, obtaining raw node embeddings $\mathcal{V}_{s/t}^{raw}$ after sampling fine-grained embeddings. Additionally, \method{} utilizes the node alignment loss $\mathcal{L}_{NR}$ to mitigate node overfitting, as proposed in previous work \cite{li2022sigma}.

\subsubsection{Contrastive Learning for Semantic Node Refinement}
Graph-based Domain Adaptive Object Detection (DAOD) methodologies commonly rely on ground-truth boxes and pseudo score maps to sample semantic nodes. Nevertheless, these nodes frequently encounter challenges stemming from uninformative backgrounds and indistinct foreground semantics, notably observable in situations such as blurred pixels resulting from overlapping objects in adverse weather conditions. Attempting to align such non-informative noises across domains typically yields no performance improvement and may even introduce confusion during cross-domain alignment.
To address this challenge, we capitalize on the observation that absent nodes from the memory bank display diminished noise levels owing to momentum update mechanisms. 
Hence, our aim is to reconstruct the characteristics of raw nodes by leveraging their categorical connections with absent nodes stored in the memory bank. To achieve this, we utilize methodologies of Graph-guided Memory Bank~\cite{li2022sigma} to produce hallucination nodes representing missing categories $\Delta^{miss}$ and $\Delta^{miss}$. Consequently, this process yields enhanced node embeddings $\mathcal{V}_{s/t}^*$ derived from the initial raw embeddings $\mathcal{V}_{s/t}^{raw}$. 
Here, $\mathcal{V}_{s/t}^{raw}$ denotes the initial raw node embeddings, while $\mathcal{V}_{s/t}^*$ signifies the enhanced embeddings.
In the process of reconstructing raw node features, a self-attention module is employed on $\mathcal{V}_{s/t}^*$ to facilitate communication between absent nodes and raw nodes. Furthermore, we observe that noise in $\mathcal{V}_{s/t}^{raw}$ compromises its low-rank characteristic, crucial for semantic tasks such as detection and recognition. Moreover, noise tends to dominate the covariance of $\mathcal{V}_{s/t}^{raw}$, adversely affecting its features.
To ameliorate node sampling and mitigate noise, we leverage the reconstructed node features $\mathcal{V}_{s/t}$ to refine $\mathcal{V}_{s/t}^{raw}$. Specifically, we utilize Implicit Feature Extraction to project $\mathcal{V}_{s/t}^{raw}$, $\mathcal{V}_{s/t}$, and $\mathcal{V}_{s/t}^*$ into implicit feature spaces $\mathcal{E}_{s/t}^{ini}$, $\mathcal{E}_{s/t}^{neg}$, and $\mathcal{E}_{s/t}^{pos}$, respectively. Subsequently, we propose a Contrastive Distribution Loss $\mathcal{L}_{CNC}$ aimed at minimizing the disparity between $\mathcal{E}_{s/t}^{ini}$ and $\mathcal{E}_{s/t}^{pos}$:
\begin{equation}
    \mathcal{L}_{CNC}=-\log \frac{exp\left(\left(f(\mathcal{E}_s^{ini},\mathcal{E}_s^{pos})+f(\mathcal{E}_t^{ini},\mathcal{E}_t^{pos})\right)/\tau\right)}{\sum_{i=1}^4\sum_{k=1}^4 1_{[k\neq i]}exp(f(z_i,z_k)/\tau)}.
\end{equation}
In the proposed contrastive distribution loss $\mathcal{L}_{CNC}$, the terms $z_i$ and $z_k$ denote embeddings derived from distinct feature spaces, encompassing $\mathcal{E}_s^{pos}$, $\mathcal{E}_t^{ini}$, $\mathcal{E}_s^{neg}$, and $\mathcal{E}_t^{neg}$. The parameter $\tau$ serves as the temperature factor, regulating the scale of similarity computations performed by the $f(·)$ function, which quantifies the similarity between embeddings $z_i$ and $z_k$. To prevent training collapse, stop gradient is applied to $\mathcal{E}_s^{pos}$, effectively attenuating noise stemming from the detector.
Unlike prior methodologies such as~\cite{li2022sigma}, which directly generate noisy nodes, our approach focuses on generating reconstructed nodes with reduced noise. Leveraging contrastive learning enhances feature extraction and node sampling within the model, culminating in a semantic-complete node set $\mathcal{V}_{s/t}$ conducive for subsequent graph modeling.

\subsubsection{Domain Alignment for Node}
The intrinsic domain-specific characteristics embedded within generated nodes influence their representation and the extraction of class-invariant features, consequently affecting semantic alignment. Research efforts such as~\cite{pan2019switchable,sun2016deep} have demonstrated that feature covariance encompasses domain-specific traits, notably color, which significantly impact representation learning. Extending this insight to graph-based representations, where nodes encapsulate comprehensive graph information including highly concentrated semantic features, becomes imperative.
To mitigate this challenge, we propose leveraging a shared attention layer for processing nodes $\mathcal{V}_{s/t}$. This layer is designed to model the categories distribution while capturing long-dependence semantic dependencies, thereby enhancing node representations. Consequently, we obtain refined node representations denoted as $\tilde{\mathcal{E}}_{s/t}$.
To achieve precise category alignment between source and target domains, inspired by ~\cite{choi2021robustnet}, we propose the integration of a Domain Alignment for Node into our framework. The Domain Alignment for Node module seeks to disentangle domain-specific elements from class-invariant features encoded within node covariance matrices. Its objective is to selectively eliminate domain-specific elements, thereby enhancing the accuracy of semantic alignment across different domains.
Initially, within the starting $n$ epochs, where $n$ is a hyper-parameter, and set to 30 as default, we gather the statistics of covariance matrices derived from $\mathcal{V}_{s/t}$. Subsequently, we compute two covariance matrices from nodes $\mathcal{V}_{s/t}$ and derive the variance matrix by computing differences between these two covariance matrices. In formula, the variance matrix $\Xi\in \mathbb{R}^{C\times C}$ is defined as:
    $\Xi=\frac{1}{N}\sum_{i=1}^N \sigma_i^2$.
The procedure involves extracting both the mean ($\mu_{\sum_i}$) and variance ($\sigma_i^2$) for each element from two covariance matrices of the $i$-th graph, thus:
\begin{equation}
    \mu_{\sum_i}=\mathbb{E}[\sum_c ([\mathcal{E}_s^i]):\sum_c([\mathcal{E}_t^i])],
\end{equation}
\begin{equation}
    \sigma_i^2=\mathbb{E}[([\sum_c ([\mathcal{E}_s^i]):\sum_c([\mathcal{E}_t^i])]-\mu_{\sum_i})^2].
\end{equation}
In the proposed method, $[:]$ denotes concatenation, and $\sum_c(\cdot)$ means the covariance matrix of the embedding of $\mathcal{E}_{s/t}^i$. Consequently, $\Xi$ comprises elements representing the variance of covariance elements between source and target domains. The matrix $\Xi$ signifies the sensitivity of the respective covariance to shifts in domains, with elements exhibiting high variance values indicative of domain-specific styles.
To isolate such elements, we employ $k$-means clustering on the strict upper triangular elements $V_{i,j}$ (where $i < j$) of the variance matrix $V$, thereby assigning elements into $k$ clusters $C = \{c_1, c_2, \cdots, c_k\}$ based on their values. Subsequently, we partition the $k$ clusters into two groups: $G_{high} = \{c_{m+1}, \cdots, c_k\}$ comprising elements with high variance values. Lastly, we propose the Node Alignment loss, which is formulated as follows:
\begin{equation}
    \mathcal{L}_{NA}=\mathbb{E}[||\sum_c([\mathcal{V}_s^i:\mathcal{V}_t^i])\odot \tilde{A}||_1],
\end{equation}
where $\tilde{A}\in\mathbb{R}^{C\times C}$ is a upper triangular matrix.

\subsection{Graph Optimization}
To attain Precision Graph Optimization, a comprehensive comprehension of domain knowledge within each source and target graph is imperative. 
Effective message propagation across graphs becomes pivotal for augmenting the model's perceptiveness towards odd node representation. Therefore, we carefully designed efficiently message-passing mechanism~\cite{yin2022dynamic,yin2023messages}:
\begin{equation}
    \mathbf{E}=Norm\{Att.(\tilde{\mathcal{E}})+\tilde{\mathcal{E}}\}.
\end{equation}
In the equation, $\tilde{\mathcal{E}}=[\tilde{\mathcal{E}_s}:\tilde{\mathcal{E}_t}]$, and $\mathbf{E}$ denotes the node set with a holistic perception of the domain. $Norm$ signifies layer normalization~\cite{ba2016layer}, and $Att.$ refers to the self-attention operator. We derive the graph nodes $\hat{\mathcal{E}}_{s/t}=\{\hat{e}_{s/t}^i\}_{i=1}^{N_{s/t}}$ of different domains following the concatenation. 
To enhance the semantic richness of the graphical representations, we utilize the  classification loss $\mathcal{L}_{node}$, implemented via a classifier $f_{cls}$ integrated with the Cross Entropy loss:
\begin{equation}
    \mathcal{L}_{node}=-\sum_{i=1}^{N_s+N_t}y_i \log \{softmax[f_{cls}(\hat{e}_{s/t}^i)]\}.
\end{equation}
In this equation, $y_i$ represents the ground-truth label for source nodes, while the pseudo label is denoted for target nodes. The process of message propagation ensures that each node undergoes enhancement based on the knowledge exchanged with other nodes, facilitating Graph Optimization.

\subsection{Framework}
During the training phase of \method{}, the comprehensive loss function $\mathcal{L}$ comprises the following components:
\begin{equation}
    \mathcal{L}=\lambda_1\mathcal{L}_{CNC}+\lambda_2\mathcal{L}_{NA}+\mathcal{L}_{GA}+\mathcal{L}_{det}+\mathcal{L}_{node},
\end{equation}
where $\mathcal{L}_{GA}$ is the alignment loss~\cite{hsu2020every}, $\mathcal{L}_{NA}$ is the alignment loss for nodes and $\mathcal{L}_{det}$ is the detection loss~\cite{tian2019fcos}. $\lambda_{1/2}$ are set 0.1 respectively to control adaptation weight.

%% file: code/5_exp.tex
\section{Experiments}
\subsection{Datasets}
Extensive experimental analyses are conducted across three distinct adaptation scenarios, adhering to the well-established paradigm of Unsupervised Domain Adaptation (UDA) as outlined in seminal literature \cite{chen2018domain,hsu2020every,munir2021ssal}. This paradigm involves training models on annotated source data while leveraging unlabeled target data for adaptation, followed by evaluation on the target dataset. Evaluation metrics, including mean Average Precision (AP) computed across various Intersection over Union (IoU) thresholds, facilitate comprehensive comparisons. Additionally, to assess adaptation efficacy under varied conditions, source-only results (SO) are juxtaposed with adaptation gains (GAIN), calculated as deviations from the source-only baseline. These comparisons offer a nuanced insight into adaptation effectiveness across different experimental settings.

\begin{table*}[t]
\caption{Results on the Cityscapes→Foggy Cityscapes adaptation scenario (\%) using VGG-16 backbone networks. We provide details regarding the utilization of additional models, such as the style-transfer model and auxiliary teacher model.}
\centering
\label{table_1}
\footnotesize
\setlength{\tabcolsep}{2.2mm}
\begin{tabular}{l|cccccccc|c|cc}
\toprule
Method  &person &rider &car &truck &bus &train &motor &bike &$AP_{0.5}$ &SO &GAIN   \\
\midrule
CFFA &34.0  &46.9  &52.1 &30.8 &43.2 &29.9 &34.7 &37.4 &38.6 &20.8 &17.8 \\ 
RPNPA &33.6  &43.8 &49.6 &32.9 &45.5 &46.0 &35.7 &36.8 &40.5 &20.8 &19.7\\ 
UMT   &33.0  &46.7 &48.6 &34.1 &56.5 &46.8 &30.4 &37.4 &41.7 &21.8 &19.9\\
MeGA   &37.7  &49.0  &52.4 &25.4 &49.2 &46.9 &34.5 &39.0 &41.8 &24.4 &17.4\\
ICCR-VDD &33.4  &44.0  &51.7 &33.9 &52.0 &34.7 &34.2 &36.8 &40.0 &22.8 &17.2  \\
KTNet    &46.4  &43.2 &60.6 &25.8 &41.2 &40.4 &30.7 &38.8 &40.9 &18.4 &22.5 \\ 
TIA  &34.8  &46.3  &49.7 &31.1 &52.1 &48.6 &37.7 &38.1 &42.3 &20.3 &22.0\\ 
MGA 	 &43.9  &49.6  &60.6 &29.6 &50.7 &39.0 &38.3 &42.8 &44.3 &25.2 &18.8 \\
FGRR &33.5  &46.4  &49.7 &28.2 &45.9 &39.7 &34.8 &38.3 &39.6 &19.3 &20.3 \\ 
TDD  &39.6  &47.5  &55.7 &33.8 &47.6 &42.1 &37.0 &41.4 &43.1 &17.3 &25.8  \\
DBGL  &33.5  &46.4  &49.7 &28.2 &45.9 &39.7 &34.8 &38.3 &39.6 &19.3 &20.3  \\ 
EPM &41.9 &38.7 &56.7 &22.6 &41.5 &26.8 &24.6 &35.5 &36.0 &18.4 &17.6\\
SSAL &45.1 &47.4 &59.4 &24.5 &50.0 &25.7 &26.0 &38.7 &39.6 &20.4 &19.2\\
SCAN++ &44.2 &43.9 &57.9 &28.2 &48.1 &51.2 &30.1 &39.5 &42.8 &18.4 &24.4\\
SIGMA &46.9 &48.4 &63.7 &27.1 &50.7 &35.9 &34.7 &41.4 &43.5 &18.4 &25.1 \\
IGG &46.2 &47.0 &61.9 &27.3 &54.1 &52.3 &30.6 &40.4 &45.0 &18.4 &26.6 \\
\midrule
\method{} &47.3 &48.2 &62.1 &30.2 &53.9 &53.4 &36.8 &42.2 &46.1 &18.2 &27.1\\
\bottomrule
\end{tabular}
\end{table*}

\begin{table*}[t]
\caption{Results on the Cityscapes→Foggy Cityscapes adaptation scenario (\%) using ResNet-50 backbone networks. We provide details regarding the utilization of additional models, such as the style-transfer model and auxiliary teacher model.}
\centering
\label{table_2}
\footnotesize
\setlength{\tabcolsep}{2.2mm}
\begin{tabular}{l|cccccccc|c|cc}
\toprule
Method  &person &rider &car &truck &bus &train &motor &bike &$AP_{0.5}$ &SO &GAIN   \\
\midrule
GPA &32.9  &46.7  &54.1 &24.7 &45.7 &41.1 &32.4 &38.7 &39.5 &22.8 &16.7 \\ 
DIDN &38.3  &44.4 &51.8 &28.7 &53.3 &34.7 &32.4 &40.4 &40.5 &28.6 &11.9\\ 
DSS   &42.9  &51.2 &53.6 &33.6 &49.2 &18.9 &36.2 &41.8 &40.9 &22.8 &18.1\\
SDA   &38.8  &45.9  &57.2 &29.9 &50.2 &51.9 &31.9 &40.9 &43.3 &22.8 &20.5\\
MTTrans &47.7  &49.9  &65.2 &25.8 &45.9 &33.8 &32.6 &46.5 &43.4 &- &-  \\
AQT    &45.2  &52.3 &64.4 &27.7 &47.9 &45.7 &35.0 &45.8 &45.5 &- &- \\ 
EPM  &39.9  &38.1  &57.3 &28.7 &50.7 &37.2 &30.2 &34.2 &39.5 &24.2 &15.3\\ 
SIGMA &44.0 &43.9 &60.3 &31.6 &50.4 &51.5 &31.7 &40.6 &44.2 &24.2 &20.0 \\
IGG &44.3 &44.8 &62.2 &35.8 &54.2 &50.7 &38.2 &38.7 &46.1 &24.2 &21.9 \\
\midrule
\method{} &45.4 &45.1 &62.4 &36.2 &53.9 &50.5 &38.8 &38.2 &46.4 &24.3 &22.3\\
\bottomrule
\end{tabular}
\end{table*}

\begin{table}[t]
\caption{Comparison of results (\%) on the Sim10K→Cityscapes (S→C) and KITTI→Cityscapes (K→C) adaptation scenarios using VGG-16 backbone architecture.}
\centering
\label{table_3}
\footnotesize
\setlength{\tabcolsep}{2.5mm}
\begin{tabular}{l|cc|cc}
\toprule
Method  &S$\rightarrow$ C &SO/GAIN &K$\rightarrow$ C &SO/GAIN   \\
\midrule
EPM &49.0  &39.8/9.2  &43.2 &34.4/8.8 \\ 
DSS &44.5  &34.7/9.8 &42.7 &34.6/8.1\\ 
MeGA   &44.8  &34.3/10.5  &43.0 &30.2/12.8\\
KTNet &50.7  &39.8/10.9  &45.6 &34.4/11.2 \\
SSAL    &51.8  &38.0/13.8 &45.6 &34.9/10.7 \\ 
SCAN++  &53.1  &39.8/13.3  &46.4 &34.4/12.0\\ 
SIGMA 	 &53.7  &39.8/13.9  &45.8 &34.4/11.4\\
IGG &58.4 &39.8/18.6 &50.4 &34.4/16.0 \\
\midrule
\method{} &59.3 &40.2/18.9 &50.8 &34.8/16.4\\
\bottomrule
\end{tabular}
\vspace{-0.3cm}
\end{table}

\begin{table}[t]
\caption{Ablation studies on the Cityscapes→Foggy Cityscapes adaptation scenario using the VGG-16 backbone architecture (\%).}
\centering
\label{table_4}
\footnotesize
\setlength{\tabcolsep}{2.5mm}
\begin{tabular}{l|cccc}
\toprule
Method  &w/o &AP &$AP_{0.5}$ &$AP_{0.75}$   \\
\midrule
Baseline &- &17.0 &35.3 &15.4\\
\midrule
\multirow{5}{*}{+NR} &CNC  &23.1  &41.9 &22.7 \\
&Attn. &23.4 &42.8 &21.6\\
&NCA &23.1 &43.5 &21.4\\
&NA &21.3 &41.5 &21.2\\
&- &24.3 &42.8 &22.7\\
\midrule
\multirow{2}{*}{\shortstack{+NR\\{+GO}}} &GNE &24.1 &43.4 &22.5\\
&- &23.4 &45.3 &22.1\\
\bottomrule
\end{tabular}
\end{table}

\begin{table*}[t]
\caption{Results on the Cityscapes→Foggy Cityscapes adaptation scenario (\%) with varying cluster $k$ and different initial epochs $n$ settings.}
\centering
\label{table_5}
\footnotesize
\setlength{\tabcolsep}{3mm}
\begin{tabular}{l|c|cccccccc|c}
\toprule
$k$  &$n$ &person &rider &car &truck &bus &train &motor &bike &$AP_{0.5}$   \\
\midrule
\multirow{3}{*}{2} &20 &43.2 &42.6 &61.9 &33.4 &49.3 &38.9 &33.5 &36.3 &42.1\\
&30 &46.4 &45.2 &62.1 &27.3 &53.9 &43.8 &30.5 &41.2 &43.4\\
&40 &45.1 &40.6 &62.3 &28.7 &50.4 &51.7 &30.5 &41.0 &43.7\\
\midrule
\multirow{3}{*}{3}  &20 &45.6 &41.8 &62.9 &29.7 &52.4 &41.3 &31.9 &42.1 &43.6\\
&30 &45.3 &43.9 &61.9 &29.7 &50.4 &46.3 &33.6 &41.7 &44.2\\
&40 &46.6 &42.8 &62.3 &24.7 &53.6 &48.7 &30.9 &41.5 &44.2\\
\midrule
\multirow{3}{*}{4} &20 &44.5 &44.6 &60.9 &35.5 &53.6 &42.1 &33.7 &41.2 &44.9\\
&30 &46.5 &47.3 &62.2 &27.5 &54.3 &52.6 &30.3 &40.6 &45.3\\
&40 &46.6 &44.4 &61.6 &25.4 &54.6 &46.3 &30.6 &42.5 &44.2\\
\midrule
\multirow{3}{*}{5} &20 &45.4 &43.4 &62.6 &30.3 &50.5 &45.7 &32.3 &40.7 &43.3\\
&30 &45.3 &41.7 &60.2 &29.6 &51.5 &55.3 &30.6 &42.3 &44.7\\
&40 &45.8 &44.2 &62.6 &27.3 &51.5 &46.2 &32.6 &40.2 &43.2\\
\bottomrule
\end{tabular}
\end{table*}

\begin{table*}[t]
\caption{Results on Cityscapes→Foggy Cityscapes (\%) with different degree $p$ of OOD on graphs in the ASF adaptor.}
\centering
\label{table_6}
\footnotesize
\setlength{\tabcolsep}{3mm}
\begin{tabular}{l|cccccccc|c}
\toprule
$p$  &person &rider &car &truck &bus &train &motor &bike &$AP_{0.5}$   \\
\midrule
0.1 &46.5 &43.9 &61.8 &29.3 &52.3 &48.1 &34.5 &41.2 &44.9\\
0.05 &46.3 &44.6 &62.1 &29.5 &51.9 &51.2 &33.5 &41.3 &45.2\\
0.025 &46.6 &47.3 &62.3 &27.5 &54.5 &52.2 &30.9 &40.4 &44.8\\
0.01 &46.3 &41.8 &61.9 &31.6 &53.3 &51.6 &34.3 &40.1 &45.3\\
\bottomrule
\end{tabular}
\end{table*}

\textbf{Cityscapes→Foggy Cityscapes.}
Cityscapes~\cite{cordts2016cityscapes} is a dataset of urban landscape images captured under dry weather conditions, annotated across eight categories. It consists of a training set with 2975 images and a validation set with 500 images. Foggy Cityscapes~\cite{sakaridis2018semantic} is a synthesized dataset derived from Cityscapes, simulating foggy weather conditions. Our investigation focuses on the adaptation from normal to foggy weather conditions, exploring the domain gap in this scenario using established methodologies from existing literature.

\textbf{Sim10k→Cityscapes.} 
Sim10k~\cite{johnson2016driving} is a simulated dataset containing 10,000 images with annotated car bounding boxes. It serves as a platform for exploring the domain gap between synthetic and real-world scenes, specifically in comparison to the Cityscapes dataset. Our study aims to understand the disparity between synthesized and real-world images, aligning with prior research in the field.

\textbf{KITTI→Cityscapes.}
KITTI~\cite{geiger2012we} comprises real-world scene images captured using various camera setups, with annotations for car categories. With a dataset size of 7,481 images, KITTI is utilized to assess the potential for bidirectional adaptation between these two datasets. Our investigation explores the feasibility of adaptation in both directions, following established methodologies in the literature.

\subsection{Implementation Details}
We utilize the VGG-16 \cite{simonyan2014very} and ResNet-50 \cite{he2016deep} pre-trained models on the ImageNet dataset \cite{recht2019imagenet} as the backbone networks for our experiments.The optimization process employs the Stochastic Gradient Descent (SGD) optimizer with a learning rate of 0.0025, a momentum of 0.99, and a weight decay of $1 \times 10^{-4}$. 
Parameters $k$ and $m$ in the domain alignment for node module are configured to 4 and 1, respectively. The degree of Out-of-Distribution of graphs $p$ is set to 0.03 as default. Additionally, the threshold parameter $\tau$ for foreground/background segmentation is set to 0.6 to ensure compliance with the activation condition of the sigmoid function. Furthermore, $\tau$ is set to 0.06 as default, aligning with employed values in \cite{lin2017focal,redmon2018yolov3}. The proposed baselines comprise the FCOS \cite{tian2019fcos} object detector, while our proposed method is trained on four NVIDIA RTX 4090 GPUs.

\subsection{Results Analysis}
\textbf{Cityscapes→Foggy Cityscapes.}
Table \ref{table_1} and \ref{table_2} showcases a comprehensive comparison between \method{} utilizing VGG-16 and ResNet-50 backbones. Impressively, employing VGG-16 and ResNet-50 backbones, \method{} achieves $AP_{0.5}$ scores of 46.1\% and 46.4\%, respectively, signifying substantial performance advancements over existing methodologies. Specifically, compared to category adaptation approaches on graphs employing the VGG-16 backbone, such as SIGMA \cite{li2022sigma} and SCAN++ \cite{li2022scan++}, \method{} demonstrates notable improvements of 3.3\% and 2.6\% in $AP_{0.5}$, respectively, underscoring its superior capability in effectively modeling graphs compared to these coarser modeling strategies. Additionally, \method{} surpasses EPM \cite{hsu2020every}, KTNet \cite{tian2021knowledge}, SSAL \cite{munir2021ssal}, and TIA \cite{zhao2022task} in $AP_{0.5}$, respectively. 
Furthermore, compared to the methods employing Deformable DETR as the detector, such as AQT \cite{huang2022aqt} and MTTrans \cite{yu2022mttrans}, \method{} achieves improvements of 0.9\% and 3.0\% in $AP_{0.5}$, respectively.

\textbf{Sim10k→Cityscapes.}
Table \ref{table_3} illustrates the adaptation results, showcasing \method{}'s state-of-the-art detection performance with an $AP_{0.5}$ of 59.3\%. This represents significant enhancements over existing methodologies. Particularly, \method{} outperforms EPM \cite{hsu2020every} (49.0\% $AP_{0.5}$), KTNet \cite{tian2021knowledge} (50.7\% $AP_{0.5}$), and SIGMA \cite{li2022sigma} (53.7\% $AP_{0.5}$) by margins of 10.3\%, 8.6\%, and 5.6\% in $AP_{0.5}$, respectively, when deployed within the same pipeline, highlighting the efficacy.

\textbf{KITTI→Cityscapes.} 
Table~\ref{table_3} shows the comparison results of proposed \method{} with baselines. From Table~\ref{table_3}, we find that the \method{} significant improvement with a 50.8\% $AP_{0.5}$ and a comparable adaptation gain of 16.4\% $AP_{0.5}$ compared to state-of-the-art methods. Notably, when compared with SSAL \cite{munir2021ssal}, SCAN++ \cite{li2022scan++}, and SIGMA \cite{li2022sigma}, our approach exhibits advantages in terms of adaptation performance.

\subsection{Ablation Studies}
Table~\ref{table_4} shows the ablation studies results.

\textbf{Semantic Sampling Transformation:} The impact of integrating the Semantic Sampling Transformation module is demonstrated in Table~\ref{table_4}, showing a significant improvement of $AP_{0.5}$ with an 8.3\% $AP_{0.5}$ gain over the baseline model. Further analysis involves dissecting each sub-component to assess its efficacy. Substituting the Contrastive Learning for Semantic Node Refinement module with the previous method results in a slight decrease of $AP_{0.5}$ due to non-informative noises. Besides, by removing the $Attn.$ layer leads to a performance decline due to coarse node representation. The exclusion of the Domain Alignment for Node limits the node optimization, affecting the semantic alignment. Notably, omitting the NA loss results in a significant drop due to the overfitting of graphs.

\textbf{Graph Optimization:} Integrating the Graph Optimization adaptor yields a notable enhancement, resulting in a remarkable 45.3\% $AP_{0.5}$, showcasing a substantial improvement over the baseline model's 35.3\% $AP_{0.5}$ with a significant gain of 10.0\% $AP_{0.5}$. The exclusion of the GNE module results in a slight decrease of 1.9\% $AP_{0.5}$, attributed to hindered node message exchange. 
Thus, the integration of each sub-component proves indispensable for \method{} to achieve its state-of-the-art results.

\subsection{Sensitivity Analysis}
To delve deeper into the efficacy of each module, we scrutinize the parameter configurations of the covariance alignment, as summarized in Table~\ref{table_5}.

\textbf{Evaluation on the covariance alignment setting:}
In Table~\ref{table_5}, we compare different settings for $k$ (the number of clusters) and $n$ (the initial epoch) for collecting covariance from nodes $\mathcal{V}_{s/t}$. It is evident that a low setting value of $k$ or $n$ ($k=1$ or $n=20$) limits the model's ability due to inadequate access to nodes information. Conversely, setting $n=30$ leads to a comprehensive performance improvement. However, collecting an excessive number of nodes covariance ($n=40$) results in a slightly inferior outcome, presumably due to the saturation of diversity in the domain-specific element.


%% file: code/6_conclusion.tex
\section{Conclusion}
In this work, we introduce a novel framework Graph Generation for Domain Adaptive Object Detection, named \method{}, addressing the pertinent challenge of utilizing graphs within the DAOD context. Our approach employs a Semantic Sampling Transformation module to ensure the creation of high-quality model graphs and accurate semantic alignment by eliminating domain-specific stylistic elements. Subsequently, we employ the Graph Optimization for domain-invariant representation learning. Through extensive experimentation across various datasets, our proposed method demonstrates superior performance compared to existing approaches in the field.